\documentclass{article}

\usepackage{microtype}
\usepackage{graphicx}
\usepackage{booktabs} % for professional tables

\usepackage{xspace}
\usepackage{subcaption}
\usepackage{algorithm}
\usepackage{amsmath}
\usepackage{hyperref}
\usepackage{makecell}
\usepackage{subcaption}
\usepackage{adjustbox}
\usepackage{enumitem}
\newenvironment{CompactItemize}
  {\begin{itemize}[noitemsep,topsep=0pt,leftmargin=*]}
  {\end{itemize}}

\usepackage[accepted]{mlsys2025}

\mlsystitlerunning{Efficient Long-context Language Model Training by Core Attention Disaggregation}

\newcommand{\sys}{DistCA\xspace}

\begin{document}

\twocolumn[
\mlsystitle{Efficient Long-context Language Model Training by \\
\emph{Core Attention Disaggregation}}

% unless you've provided the [accepted] option to the mlsys2025


\mlsyssetsymbol{equal}{*}
\mlsyssetsymbol{workdoneucsd}{\textdagger}   % uses † for work done at UCSD

\begin{mlsysauthorlist}
\mlsysauthor{Yonghao Zhuang}{equal,cmu}
\mlsysauthor{Junda Chen}{equal,ucsd}
\mlsysauthor{Bo Pang}{workdoneucsd,ucsd}
\mlsysauthor{Yi Gu}{ucsd,mbz}
\mlsysauthor{Yibo Zhu}{stepfun}
\mlsysauthor{Yimin Jiang}{stepfun}\\
\mlsysauthor{Ion Stoica}{ucb}
\mlsysauthor{Eric Xing}{cmu,mbz}
\mlsysauthor{Hao Zhang}{ucsd}
\end{mlsysauthorlist}

\mlsysaffiliation{cmu}{Carnegie Mellon University}
\mlsysaffiliation{ucsd}{UC SanDiego}
% \mlsysaffiliation{gt}{Georgia Institute of Technology}
\mlsysaffiliation{stepfun}{StepFun}
\mlsysaffiliation{mbz}{MBZUAI}
\mlsysaffiliation{ucb}{UC Berkeley}

\mlsyscorrespondingauthor{Eric Xing}{epxing@andrew.cmu.edu}
\mlsyscorrespondingauthor{Hao Zhang}{haz094@ucsd.edu}

\mlsyskeywords{Machine Learning, MLSys}

\vskip 0.3in

% 
% Abstract Original
% 
\begin{abstract}
We present core attention disaggregation (CAD), a technique that improves long-context LLM training by disaggregating the core attention (CA) -- the parameter-free $\mathrm{softmax}(\mathbf{QK}^{\top})\mathbf{V}$ computation -- and schedules it on an independent pool of resources. Existing systems co-locate core attention with other components. At long context, the quadratic growth of CA computation and near-linear growth of the rest create load imbalance -- hence stragglers across data and pipeline groups.
CAD is enabled by two key observations: (i) \emph{statelessness}: CA has no trainable parameters and minimal transient state, so balancing reduces to scheduling compute-bound tasks; and (ii) \emph{composability}: modern attention kernels sustain high utilization on fused batches of arbitrary-length token-level shards. 
CAD dynamically partitions the core attention computation into token-level tasks (CA-tasks), and dispatches them to a pool of devices specialized for CA computation (attention servers). It then rebatches CA-tasks to equalize CA compute across attention servers without loss of kernel efficiency.
We have implemented CAD in a system called \sys with a ping-pong scheme to completely overlap communication with compute, and in-place attention servers to improve memory utilization. On up to 512 H200 GPUs and 512K context length, \sys improves end‑to‑end training throughput by up to 1.35$\times$, eliminates DP/PP stragglers, and maintains near‑perfect compute and memory balance.
\end{abstract}
]



\printAffiliationsAndNotice{
\mlsysEqualContribution
    \textdagger\;Work done at UCSD (internship)
} % otherwise use the standard text.

\section{Introduction}

\begin{figure*}[t!]
    \centering
    \begin{adjustbox}{valign=T, minipage={0.75\linewidth}}
    \begin{minipage}[b]{0.49\linewidth}
        \centering
        \includegraphics[width=\linewidth]{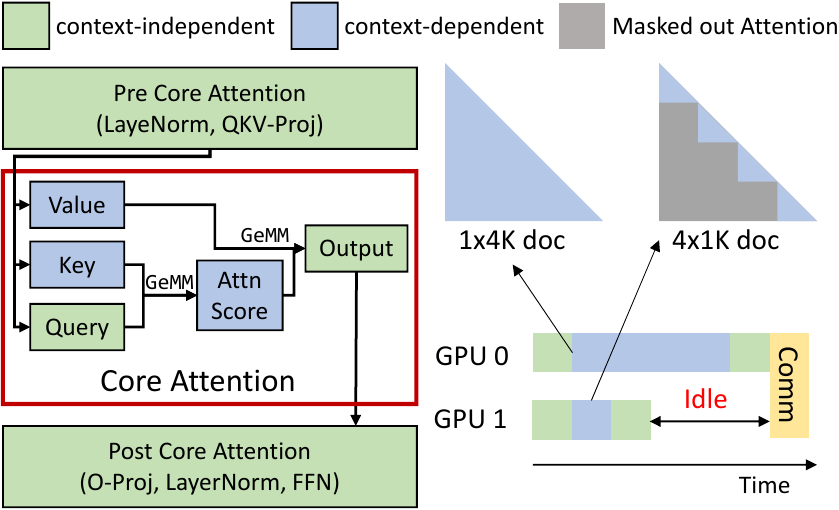}
        \caption{Transformer and its workload imbalance caused by core attention.}
        \label{fig:transformer-layer}
    \end{minipage}
    \hfill
    \begin{minipage}[b]{0.49\linewidth}
        \centering
        \includegraphics[width=\linewidth]{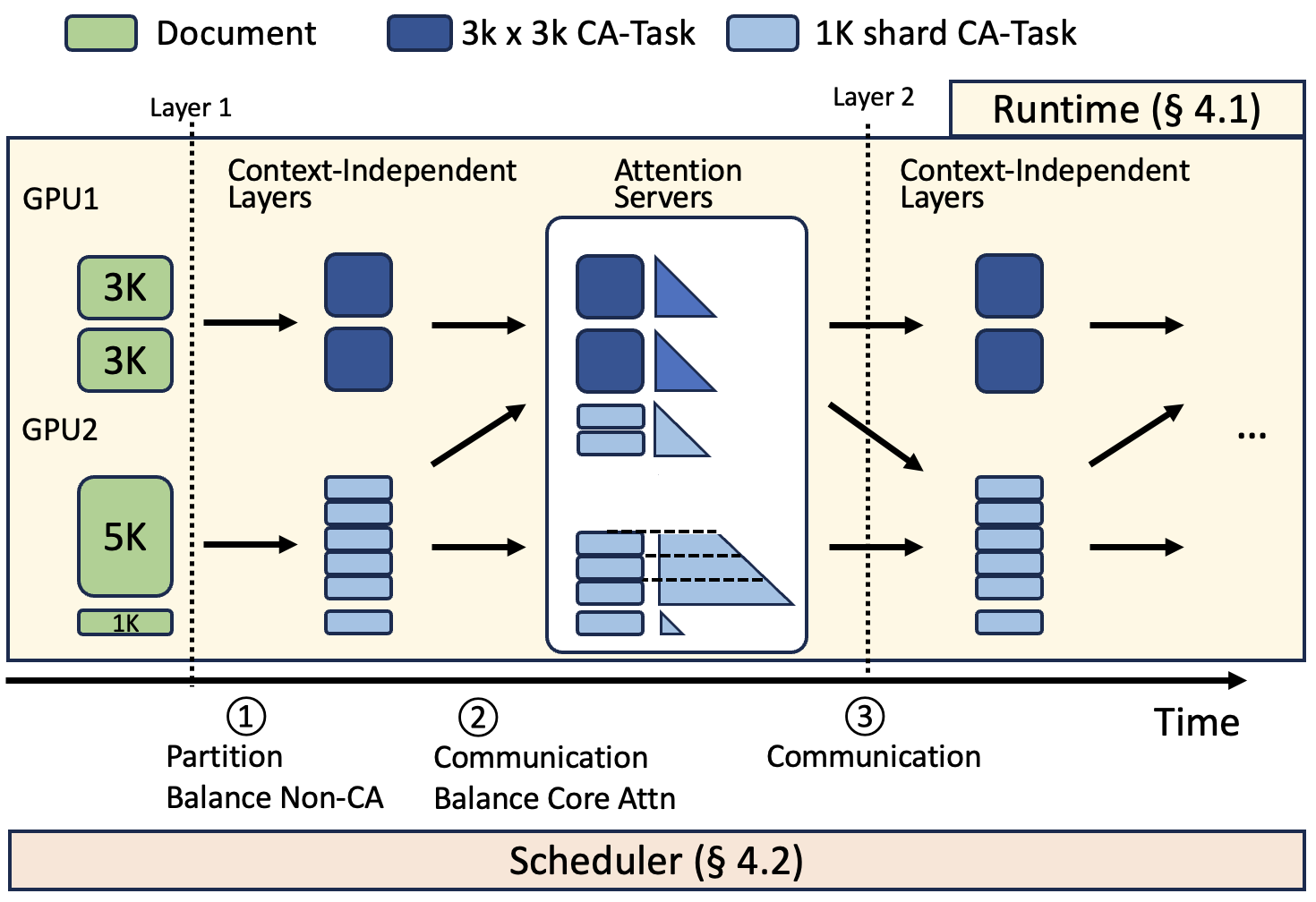}
        \vskip -0.5em
        \caption{\sys architecture.}
        \label{fig:caaas-workflow}
    \end{minipage}
    \end{adjustbox}
    \hfill
    \begin{adjustbox}{valign=T, minipage={0.24\linewidth}}
        \centering
        \small
        \vskip -1em
        \captionof{table}{Compute and Memory for components in LLMs. $l$ is the document's number of tokens.}
        \hfill
        \vskip 0.3em
        \begin{tabular}{c|c|c}
            & Memory & Compute \\
            \hline
            CA & 0 & $O(l^2)$ \\
            Linear & $O(l)$ & $O(l)$\\
            MISC & $O(l)$ & $\approx 0$
        \end{tabular}
        \label{tab:flops-memory-analysis}
        \newline\newline
        \raggedright
        \footnotesize
        CA: core attention layer;
        
        Linear: FFN, qkvo-proj
        % Linear: FFN, qkvo-proj, o-proj

        MISC: layer norm, dropout, ...
    \end{adjustbox}
    \vskip -1em
\end{figure*}

Recent large language model (LLM) applications show a steadily increasing demand for processing longer contexts. For instance, reasoning workloads must generate long chain-of-thoughts to yield accurate answers~\cite{guo2025deepseek}; coding agents operate over multi-file repositories~\cite{liu2024repobench}. To reliably support these use cases, LLMs must operate with contexts of 100K - 1M tokens at inference.

Equipping an LLM with long-context capability typically requires training on datasets that include long documents in addition to ordinary (short) documents~\cite{gao2024train}. 
A standard approach to batch documents of variable length is \emph{document packing}~\cite{rae2021scaling,wang2024packing}: concatenate multiple documents into a fixed-size chunk and apply an attention mask to block cross-document attention. 
While document packing improves throughput, it leads to variance in the attention compute per chunk -- therefore \emph{load imbalance} -- across different chunks~\cite{lin2025understanding}. The root cause is that self-attention compute in transformers grows quadratically with sequence length, whereas the rest scales approximately linearly. Consequently, two chunks with the same total tokens can incur very different attention FLOPs depending on how documents are distributed. For example, in \autoref{fig:transformer-layer}, a chunk with a single 4K-token document requires roughly 4x the attention FLOPs of a chunk packed with four 1K-token documents, even though both contain 4K tokens.

This imbalance manifests as stragglers (\autoref{fig:transformer-layer}) in large-scale distributed training in two ways~\citep{wang2025wlb,lin2025understanding}. First, in data parallelism (DP), replicas process different chunks and synchronize at the gradient barrier; the replica with the largest attention workload stalls the others. Second, in pipeline parallelism (PP)~\citep{huang2019gpipe,megatron-lm}, pipeline stages operate on different microbatches concurrently; a microbatch with a larger attention workload makes its current stage a straggler, creating pipeline bubbles which stall the entire pipeline. In hybrid DP and PP, these effects compound, and slowdowns of 1.34-1.44x~\citep{wang2025wlb,lin2025understanding} have been reported even under modest context lengths.

One remedy is to equalize compute by assigning more tokens to devices processing shorter documents~\cite{wang2025wlb}. This balances compute but unbalances memory, as activation footprints grow with total tokens. In the ``4$\times$1K vs. 1$\times$4K'' example: matching the 1$\times$4K attention FLOPs requires 12 more 1K documents on the 4$\times$1K device, raising its activation memory by $3\times$. Another remedy, context parallelism (CP)~\citep{liu2023ring,jacobs2023deepspeed}, shards each document along the sequence dimension and equally distributes shards across DP replicas, which balances compute and memory, but at the cost of extra communication of KV states. Unfortunately, it cannot mitigate PP stragglers: pipeline stages hold disjoint model layers and cannot jointly compute shards of the same document; straggler microbatches still generate bubbles that stall other stages.

Fundamentally, the imbalance stems from mismatched complexity between attention and the rest of the model computations, as illustrated in Table~\ref{tab:flops-memory-analysis}. When attention and other layers are colocated and scaled together, the mismatch grows with model scale and context length, leading to severe load imbalance.

This naturally leads us to disaggregate attention from the rest of the model, so the attention and non-attention components can be scaled \emph{independently} to balance workload across devices.
A challenge, however, is that disaggregation introduces \emph{per‑layer} transfers of inputs and outputs across the attention boundary. At first glance, this communication is prohibitive and could negate its benefits. Surprisingly, we find the opposite: precisely isolating the \emph{core attention} (CA)---the weightless $\mathrm{softmax}(\mathbf{QK}^{\top})\mathbf{V}$ computation (\autoref{fig:transformer-layer})---and applying targeted overlap and placement optimizations allows the communication to be effectively hidden in today's long-context training workloads.

Specifically, we identify two key characteristics of core attention that makes disaggregation effective. 
First, \emph{statelessness}: CA contains no trainable parameters and stores only small per-row softmax statistics; balancing CA hence reduces to scheduling compute-bound tasks. 
Second, \emph{composability}: 
CA can be partitioned at token granularity into arbitrary-length shards. 
Each shard, given its target tokens' $\mathbf{Q}$ and context tokens' $\mathbf{K}$, $\mathbf{V}$, 
independently computes the output. 
Shards from different documents (i.e., DP replicas or pipeline stages) can be re-batched into a single high-occupancy kernel. In modern attention kernels (e.g., Flash Attention), throughput primarily depends on the aggregate tokens in the fused call rather than their document of origin. This allows partitioning and recombining shards to equalize CA compute, instead of committing to the uniform splits typically assumed by CP.

These observations lead to \emph{core attention disaggregation} (CAD), which disaggregates CA and schedules it independently on a pool of resources specialized for CA computation, named \emph{attention servers}. Attention servers accept core attention tasks---the CA computation of \emph{arbitrarily partitioned} document shards---as compute requests. 
The runtime dynamically batches CA tasks into large fused kernel calls and schedule them to any attention server. 
Hence, CAD allows to achieve \emph{near-perfect} load balancing across all attention servers, while avoiding memory imbalance. Unlike CP, CAD fully decouples core attention from the parallelisms used for the rest of the model, hence eliminating stragglers in both DP and PP.

Implementing CAD raises two practical challenges.
First, CA is compute-intensive but memory-light; dedicating devices as attention servers might underutilize memory. 
We therefore time-share GPUs between attention servers~\cite{li2014communication} and the rest of model computation to keep both compute and memory utilization high (\S\ref{sec:system-features}).
Second, dispatching token-level core attention tasks adds communication overhead; we design a ping-pong execution scheme (\S\ref{sec:system-features}) to overlap communication with computation between two alternating batches. We also develop a scheduler (\S\ref{sec:as-algo}) that optimizes document sharding to strike a balance between balancing workloads while minimize communication. 

We implement CAD in a system called \sys, and evaluate it at up to 512 H200 GPUs for workloads with up to 512k context length. Compared to existing systems, \sys improves end-to-end throughput by up to 1.35x, and achieves near-linear weak scaling. Our studies show that the communication caused by CAD can be fully hidden, and attention can be balanced near-perfectly. 

In summary, this paper makes three key contributions: 

\begin{CompactItemize}
 \item Propose core attention disaggregation to address the load imbalance in long-context training of LLMs.
\item Implement \sys with three key optimizations: in-place GPU time sharing, ping-pong overlap, and a workload balanced scheduler.
\item Conduct comprehensive evaluation of CAD on large-scale realistic training workloads.
\end{CompactItemize}

\section{Background}

\subsection{LLM Architecture}
\label{sec:background-llm-arch-and-parallel}

LLMs adopt the Transformer architecture, which consists of a stack of identical Transformer layers. As shown in \autoref{fig:transformer-layer}, we distinguish two types of layers.

\noindent \textbf{Context-independent layers} include QKV-projection, output-projection, feed-forward network (FFN), layernorm, and position embedding. 
These layers operate token‑wise: the output for each token depends only on its own hidden state, so both compute and activation memory scale approximately linearly with the number of tokens. The dominant cost arises from the linear (GEMM) operators.

\noindent \textbf{Context-dependent layers} include only core attention (CA). In this paper, we deliberately distinguish \emph{core attention} from \emph{attention}; the latter refers to the composite of QKV-projection, output-projection, layernorm, and core attention in most literature.
Given queries ($\mathbf{Q}$), keys ($\mathbf{K}$), and values ($\mathbf{V}$), CA computes attention scores $\mathbf{P} = \mathrm{softmax}(\mathbf{QK}^{\top})$ (with masking as needed) and outputs $\mathbf{O} = \mathbf{P} \times \mathbf{V}$. We emphasize that CA has no trainable parameters (nor gradients) and its per-token computation does not require intermediate outputs from other tokens' CA, but only their $\mathbf{K}, \mathbf{V}$ vectors. 
The CA compute process is plotted in \autoref{fig:transformer-layer}. 

At training, storing $\mathbf{P}$ is memory-prohibitive due to its quadratic complexity. Modern IO-aware attention kernels~\cite{dao2022flashattention} avoid materializing $\mathbf{P}$ and recompute it during backward. Hence, CA also generates an negligible amount of intermediate states, rendering it stateless.

\subsection{LLM Training Parallelization}
\label{sec:parallelization}
Modern LLM training combines four parallelisms -- data, tensor, context, and pipeline parallelisms -- to scale.

In data parallelism (DP), replicas process different batches and meet at a gradient synchronization barrier. Any replica that receives a batch with a larger attention workload (e.g., from packed data) becomes a straggler and stalls the others. 

Tensor parallelism (TP) shards model layers at the cost of per-layer communication, which is not affordable when scaling beyond one node (TP size $ > 8$). 
TP balances memory and compute across TP shards, as attention is sharded along the head dimension and all devices process exactly the same data batch~\cite{megatron-lm}.

Context parallelism (CP) shards data along the sequence dimension across multiple GPUs. 
Context-independent layers run independently on each shard, yet core attention requires an all-gather (AG) to collect token states of its context to compute the attention of the local shard. Under a causal mask, earlier tokens in the sequence performs less attention computation than later tokens, so naively slicing the sequence creates load imbalance. Recent work~\cite{grattafiori2024llama} mitigates this with a head-tail shard assignment: partitions the sequence into $2 \times \mbox{CP}$ shards and assign rank $i$ both the $i$-th and the $(2 \times \mbox{CP} - 1 - i)$-th shard. 
Further, when document packing is used, \emph{per-document CP} shards each document rather than the entire concatenated chunk to balance workloads, which is discussed in detail in \S\ref{sec:imbalance-motivating-example}.

Pipeline parallelism (PP) partitions the model's layers into stages, and splits the input data into microbatches, which flow through each stage concurrently; synchronization occurs when activations are passed between two pipeline stages. To minimize pipeline bubbles, each stage shall ideally have a similar compute workload~\cite{zheng2022alpa}. 
However, because different microbatches may contain chunks packed with varying lengths of documents, the processing time of each stage become imbalanced, leading to bubbles that propagate downstream that idle all subsequent stages. 
This imbalance cannot be mitigated by globally resharding the data, as each pipeline stage processes a disjoint set of model layers.

% \section{Method}
\section{Challenge and Motivation}
This section formalizes the problem of compute and memory imbalance under document packing, explains why existing remedies cannot adequately balance both, and motivates core attention disaggregation.

\subsection{Load Imbalance in Compute and Memory}
\label{sec:workload_imbalance_problem}

Let the compute of an $l$-token document be $\mbox{FLOPs}(l) = \alpha l^2 + \beta l$, where $\alpha,\beta$ are constants related to LLM architectures (e.g.,  hidden size). $\alpha l^2$ represents the core attention FLOPs; $\beta l$ aggregates FLOPs of context-independent layers. 
Let activation memory be $\mbox{M}(l) = \gamma l$, since modern IO-aware attention kernels avoid storing $\mathbf{P}$ in forward and recompute it during backward; activations saved for backward are therefore dominated by context-independent layers.

For a microbatch of documents with lengths $\{l_i\}_{i=1}^n$ packed together, the total compute is $\alpha\sum_{i=1}^nl_i^2+\beta \sum_i^nl_i$, while the total activation memory is $\gamma\sum_{i=1}^nl_i$.
To make two microbatches -- of lengths $\{l_i\}_{i=1}^n$ and $\{l'_j\}_{j=1}^m$ -- balanced in both compute and memory, both conditions must hold: $\sum_i^nl_i=\sum_j^ml_j'$ and $\sum_i^nl^2_i=\sum_j^ml_j'^2$, which is difficult to satisfy in practice. 
Existing methods typically target one condition at a time. For example, fixed-size packing equalizes memory ($\sum_i^nl_i=\sum_j^ml_j'$) by packing the same number of total tokens in both chunks, but leaves attention compute unequal due to the quadratic term.

Two remedies, variable-length data chunk and per-document context parallelism, adjust the document packing scheme or sequence sharding to approximate these conditions~\cite{wang2025wlb}, respectively. Next, we quantitatively analyze them and show their inadequacies. 

\subsection{Problems of Existing Methods}
\label{sec:imbalance-motivating-example}

\begin{figure}[t]
    \centering
    % Upper row: ag-overhead (left) + memory_breakdown (right)
    \begin{subfigure}[t]{0.48\linewidth}
        \centering
        \includegraphics[width=\linewidth]{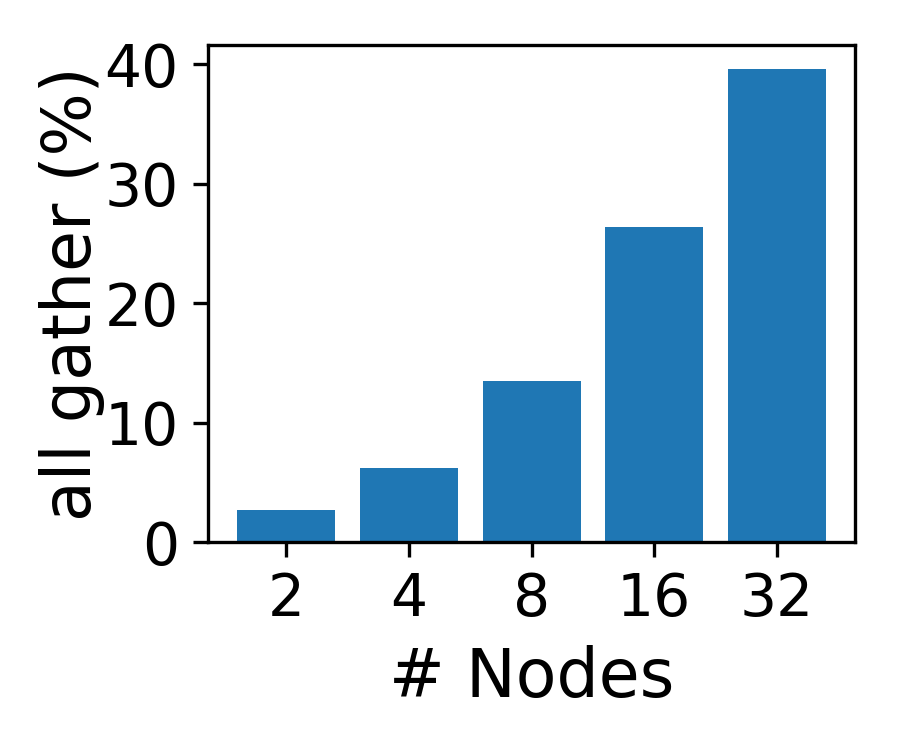}
        \caption{all-gather latency(\%)}
        \label{fig:ag-overhead}
    \end{subfigure}\hfill
    \begin{subfigure}[t]{0.48\linewidth}
        \centering
        \includegraphics[width=\linewidth]{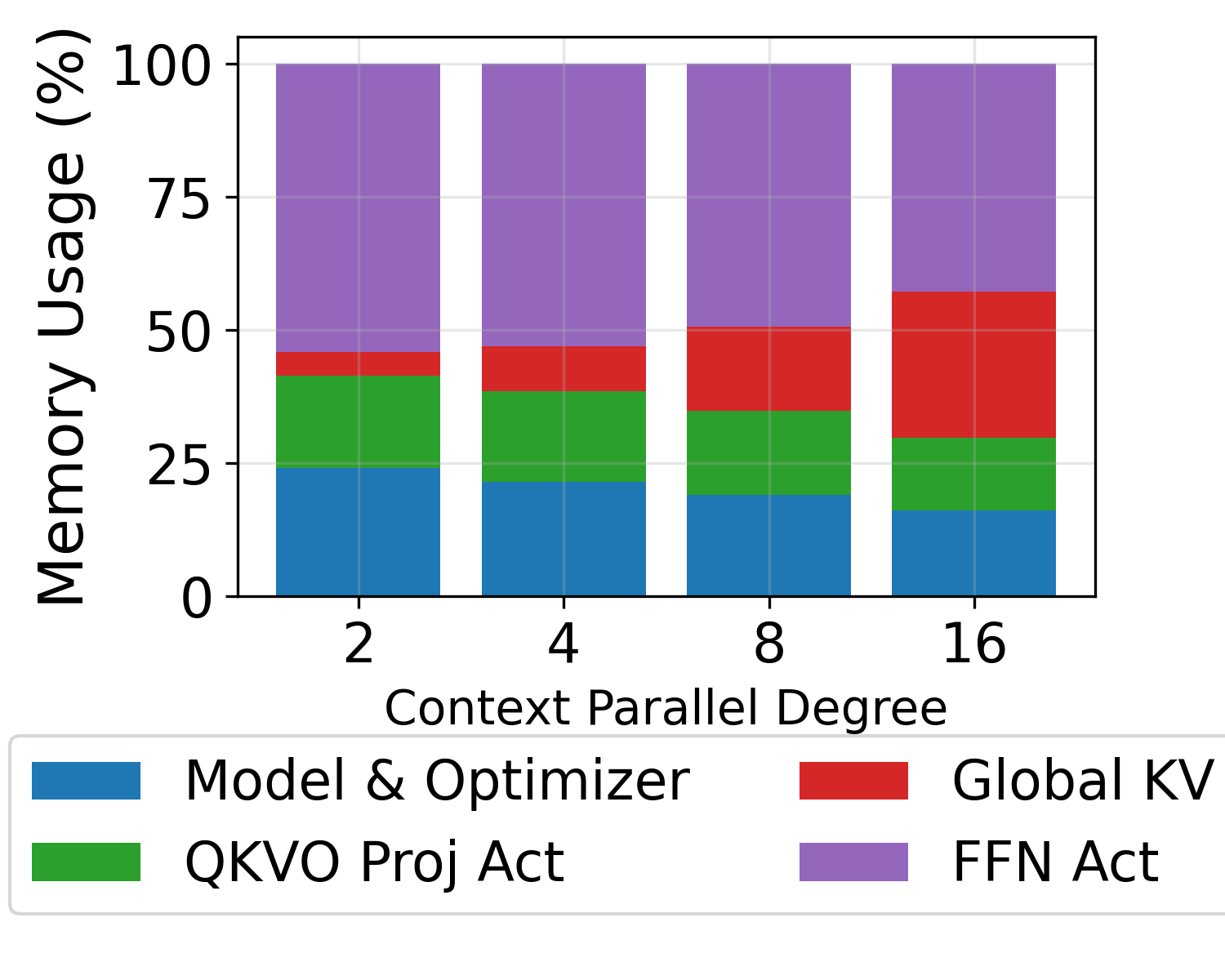}
        \caption{Memory breakdown (\%)}
        \label{fig:mem-breakdown}
    \end{subfigure}
    \caption{Latency and memory breakdown for all-gather in Context Parallel, Llama-8B. Document lengths are all 32k.}
    \label{fig:ag-scaling-behavior}
\end{figure}

\begin{figure}[t]
    \centering
    \begin{subfigure}[b]{0.45\linewidth}
        \centering
        \includegraphics[width=\linewidth]{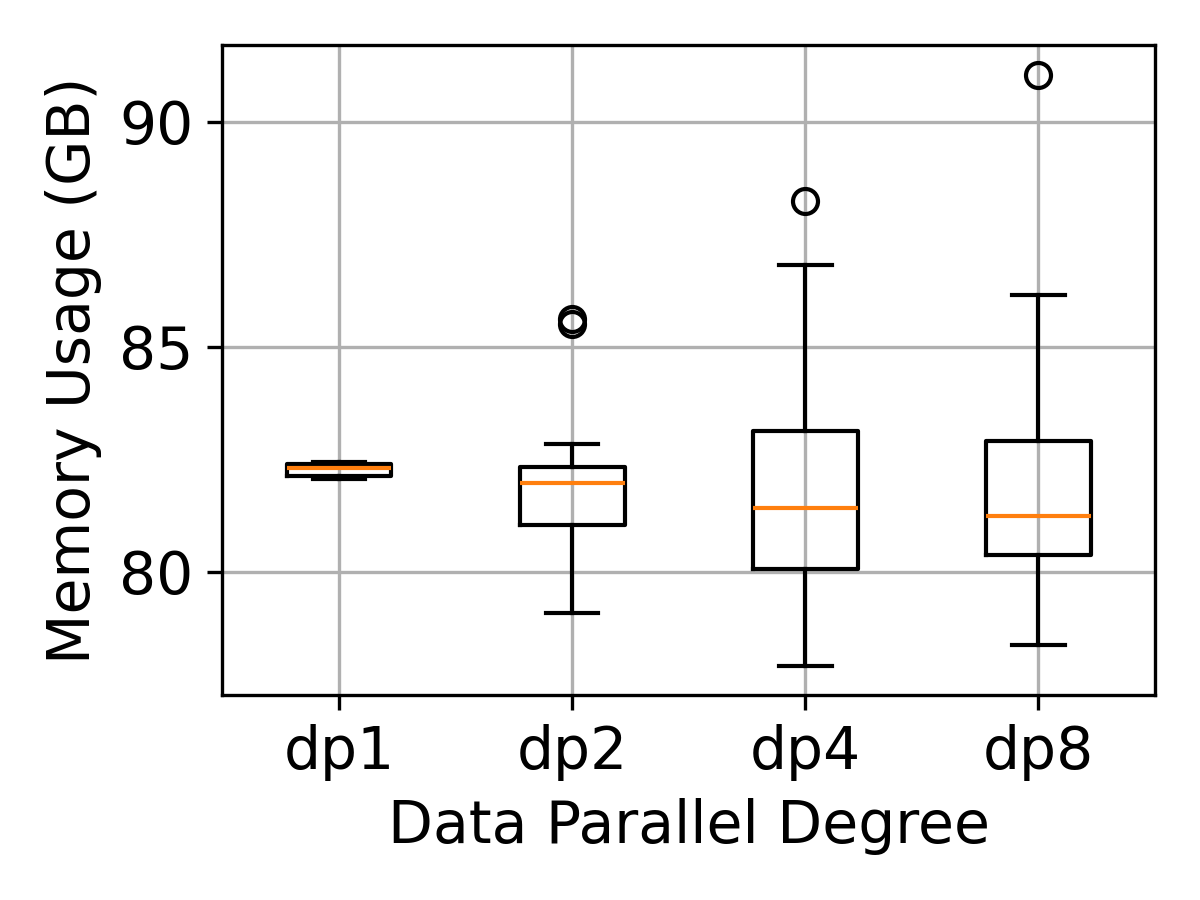}
        \caption{Memory divergence.}
        \label{fig:mem-var-len-data-chunk}
    \end{subfigure}
    \hfill
    \begin{subfigure}[b]{0.45\linewidth}
        \centering
        \includegraphics[width=\linewidth]{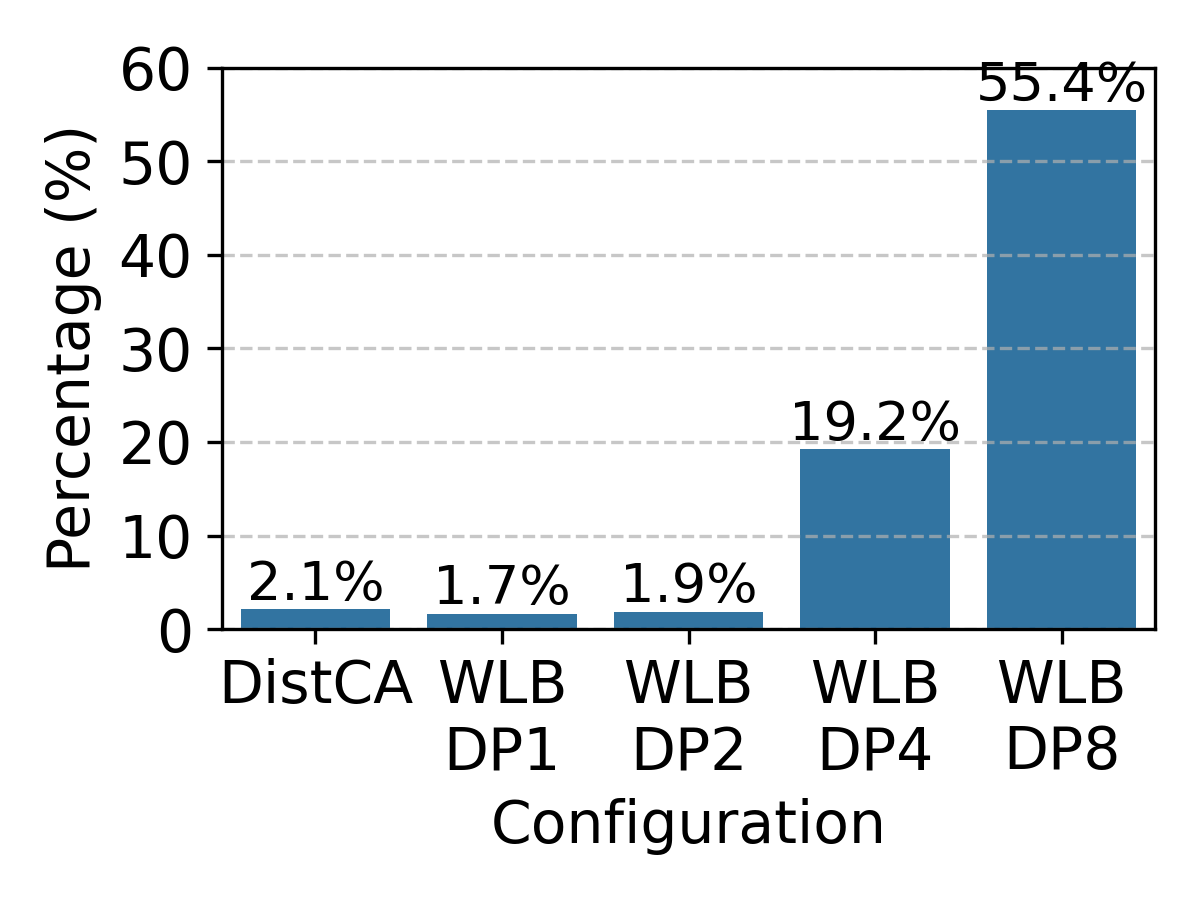}
        \caption{Divergence of Attention Computation.}
        \label{fig:ca-imbalance-compare}
    \end{subfigure}
    \caption{Throughput and memory divergence for variable-length data chunks, under different parallelism for 512K-token data chunks on 8B model.}
    \vskip -1em
\end{figure}

\noindent \textbf{Variable-length data chunk.}
This approach redistributes documents across microbatches to equalize $\sum_i^nl^2_i=\sum_j^ml_j'^2$ (FLOPs), i.e., it tries to move several documents from a microbatch with more attention compute to compensate other microbatches. However, token counts $\sum l$ then diverge across microbatches, inflating activation memory on some ranks. 
Figure~\ref{fig:mem-var-len-data-chunk} shows per-microbatch memory divergence growing with DP size (fixing TP=8), when we scale global batch size proportionally with the number of nodes to keep memory fully utilized; with a 512K maximum length, achieving compute balance requires 1.08 - 1.17x more activation memory on some ranks. 

Worse still, as sequence length grows, the method hits the memory cap, where simply moving sequences fails to fully equalize attention compute due to memory constraints. 
The resulting straggler effect is visible in Figure \ref{fig:ca-imbalance-compare}, where we quantify how much GPU compute is underutilized due to attention imbalance by measuring the percentage of average idle time to average iteration time; the idle fraction rises to 19\% at DP=4 and 55\% at DP=8 for a 512K-length workload. In such regimes, variable-length chunking is fundamentally constrained by memory pressure and becomes ineffective.

\noindent \textbf{Per-document context parallelism.}
For a chunk packed with documents $l_1,l_2\dots l_n$, evenly slicing the concatenated chunk yields imbalance because early tokens perform less work under causal masking (even with ``head-tail'' paring in  \S\ref{sec:parallelization}). Per-document CP instead \emph{partitions each document} into $2c$ shards ($c$ is the CP degree) and assigns to rank $i$ the $i$-th and $(2c - 1 - i)$-th shard of each document. Each CP rank hence processes an equal share of every document -- balancing both compute and memory. Despite this, we find that per-document CP encounters three bottlenecks at scale.

\begin{figure}[t]
  \centering
  \begin{minipage}[t]{0.48\linewidth}
    \centering
    \includegraphics[width=\linewidth]{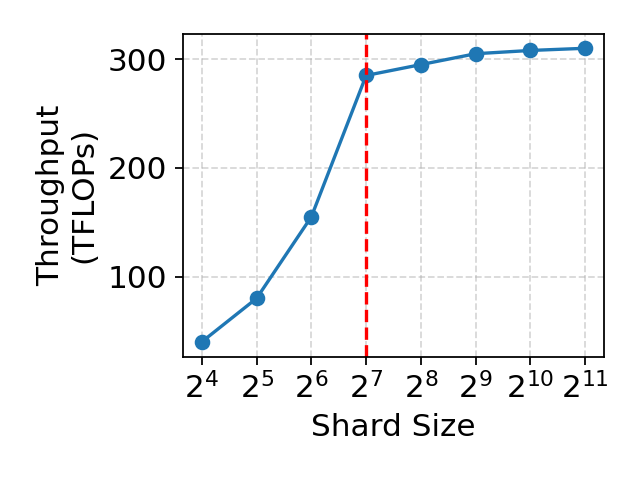}
    \caption{Throughput of core attention.}
    \label{fig:fa-throughput}
  \end{minipage}
  \hfill
  \begin{minipage}[t]{0.48\linewidth}
    \centering
    \includegraphics[width=\linewidth]{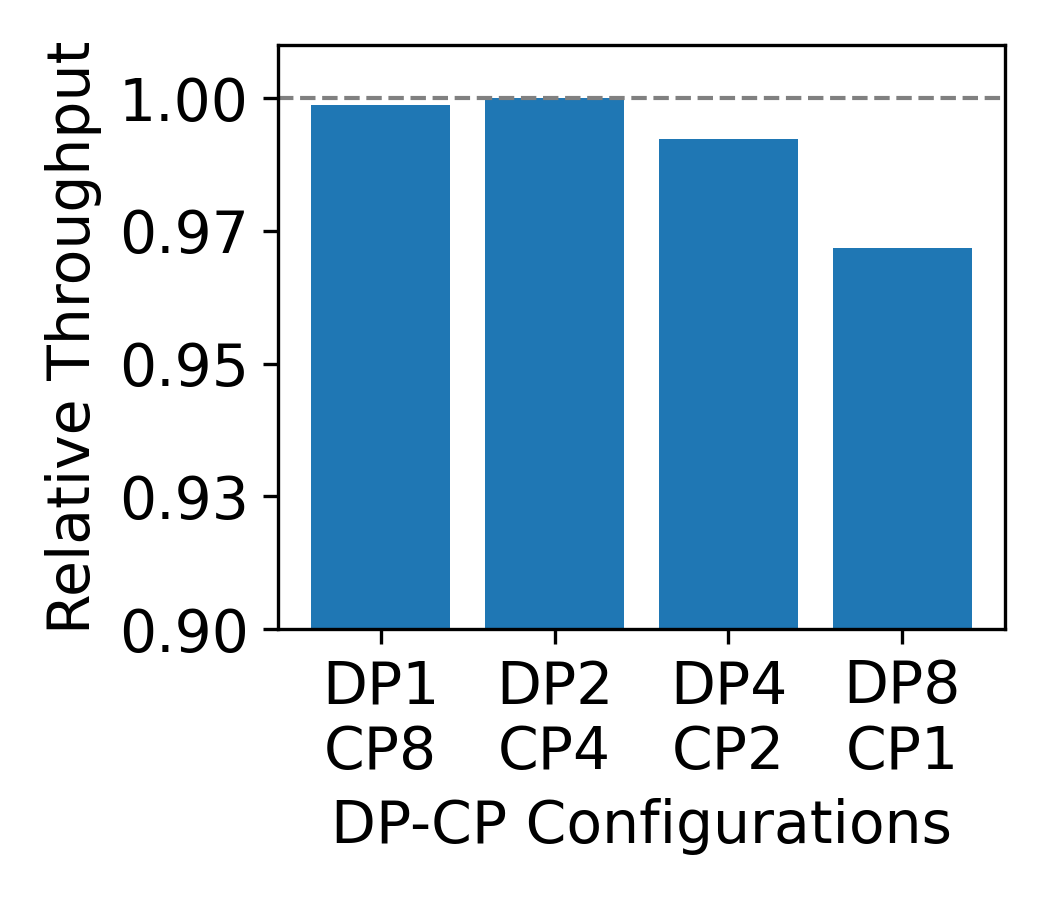}
    \caption{Throughput when applying varaible-length data chunk and per-doc CP.}
    \label{fig:optimal-wlbllm-cp-degree}
  \end{minipage}
\end{figure}

First, for ordinarily short documents, per-document CP still partitions them and results in tiny shards. 
These tiny shards may underfill attention tiles and lower arithmetic intensity~\cite{wang2025wlb}. \autoref{fig:fa-throughput} shows that the kernel's throughput drops significantly for document shorter than 128 tokens. 
Second, per-document CP reduces the per-rank compute by $1/c$ but requires all-gathering KV states, whose cost is linear to the global number of tokens. As shown in Figure~\ref{fig:ag-overhead}, as we scale the CP degree, the all-gather latency share rises from merely 3\% on 2 nodes to nearly 40\% on 32 nodes. 
Finally, the last CP rank must store the entire document's aggregated KV states for backward, causing a memory pressure. The memory overhead further increases as CP scales. As shown in Figure~\ref{fig:mem-breakdown}, the KV memory fraction grows from 3\% at 2 nodes to almost 30\% at 16 nodes. 
These bottlenecks fundamentally limit the scalability of per-document CP.

\noindent \textbf{Combination of both.} Systems that combine these techniques  inherit the drawbacks of each as scale grows. Figure~\ref{fig:optimal-wlbllm-cp-degree} shows an experiment on a 64-GPU 512K-token workload. 
Scaling the CP degree reduces imbalance, but also decreases throughput and risks OOM as batch size or number of nodes increases. On the other hand, increasing DP causes severe load imbalance and suboptimal throughput. This trade-off becomes more acute with larger scale and longer context.

\subsection{Motivation for Core Attention Disaggregation}
\label{sec:CA-disagg}

The balancing conditions in \S\ref{sec:workload_imbalance_problem} suggest a natural boundary: separate CA's quadratic term $\alpha \sum l_i^2$ from the linear terms $\beta \sum l_i$ and $\gamma \sum l_i$ associated with context-independent layers. Once CA is decoupled, we can: (1) independently schedule and equalize CA compute across a shared pool of devices, i.e. \emph{attention servers}, without concerning memory, as CA is stateless; (2) independently balance the compute and memory of context-independent layers, which are both linear with tokens. Making this practical hinges on two questions: \textbf{(Q1)} how to balance CA compute? \textbf{(Q2)} can we hide the communication caused by disaggregation? We next explain two observations that make both possible.

\noindent \textbf{Divisibility and kernel composability.} We observe that core attention computation is divisible at the token granularity. Given $\mathbf{Q}$ for target tokens and $\mathbf{K, V}$ for their context, each shard computes independently. In modern attention kernels, each GPU thread block is assigned a tile of the core attention computation. The kernel can sustain high MFU on variable-length fused sequence, provided its size is larger than this tile. In other words, kernel throughput depends primarily on the aggregate tokens in a fused call, not on their document of origin. 
This implies, in practice, documents can be arbitrarily sharded then recombined into a single high‑occupancy CA kernel (with proper masking) without hurting kernel efficiency.

We corroborate this by profiling FA2 on a 32K-token chunk packed of document shards of varying lengths and context sizes. 
Within each chunk, we fix the shard length while randomly sampling the context size for each shard; we sample multiple chunks to measure the average throughput. Results in \autoref{fig:fa-throughput} show a high throughput as long as each document shard has more than 128 tokens, which is the kernel tile size set by FA2. Shards shorter than 128 tokens are padded, which waste compute of their assigned thread blocks.
This characteristic enables us to solve \textbf{Q1} by arbitrarily partitioning any document into shards with a multiple of 128 tokens and recombine them and issue as a single attention kernel, which balances compute without compromising kernel efficiency.

% \begin{figure}[t]
%     \centering
%     \includegraphics[width=0.9\linewidth]{figures/comm_opt.pdf}
%     \caption{Communication in CP (top) vs. \sys (bottom). CP performs an all-gather operation, incurring communication of unused tensors, while \sys only transmits required tensors.}
%     \label{fig:comm-opt}
% \end{figure}

\noindent \textbf{Communication cost.} 
Like CP, disaggregating CA also requires communicating $\mathbf{K, V}$ of context tokens. 
We show that this communication has a low overhead for four reasons. 
First, due to casual mask, the $\mathbf{K, V}$ of later shards are not needed by earlier shards. 
We use an all-to-all operation to send only the required shards; 
Second, when scheduling shards onto devices, the more communication intense shards -- the later shards of a document that have more context tokens -- can be dispatched on different devices to avoid a straggler in the all-to-all; 
Third, thanks to the flexibility of arbitrary sharding, we can further optimize the sharding schemes to minimize communication, without being restricted by uniform sequence splits or a fixed CP degree. We develop this algorithm in detail in \S\ref{sec:as-algo}; 
Finally, communication sending the input of a batch's core attention can be overlapped with the computation of another batch; we realized this as a ping-pong execution scheme in \S\ref{sec:system-features}.

We formally estimate the communication overhead in~\autoref{appendix:upper-bound-ca-cp-degree}, 
and show that with a Llama-3-34B's configuration and InfiniBand bandwidth, 
documents can be partitioned into up to 31 shards and 
distributed across different devices without incurring communication overhead, 
and for larger models, this upper bound even increases.

\section{Method}
\label{sec:method}
We implement core attention disaggregation in a system named \sys. \sys has two main components: a runtime system and a workload scheduler as shown in \autoref{fig:caaas-workflow}. 
The runtime (\S\ref{sec:system-features}) 
alternate between executing the context-independent layers and the core attention of each transformer layer, inserting the necessary communication between the GPUs responsible for these two parts. The workload scheduler (\S\ref{sec:as-algo}) takes a batch of documents as input, and determines how to shard each document and where to place each shard across devices.

\subsection{Runtime}
\label{sec:system-features}
\begin{figure}
    \centering
    \includegraphics[width=1.02\linewidth]{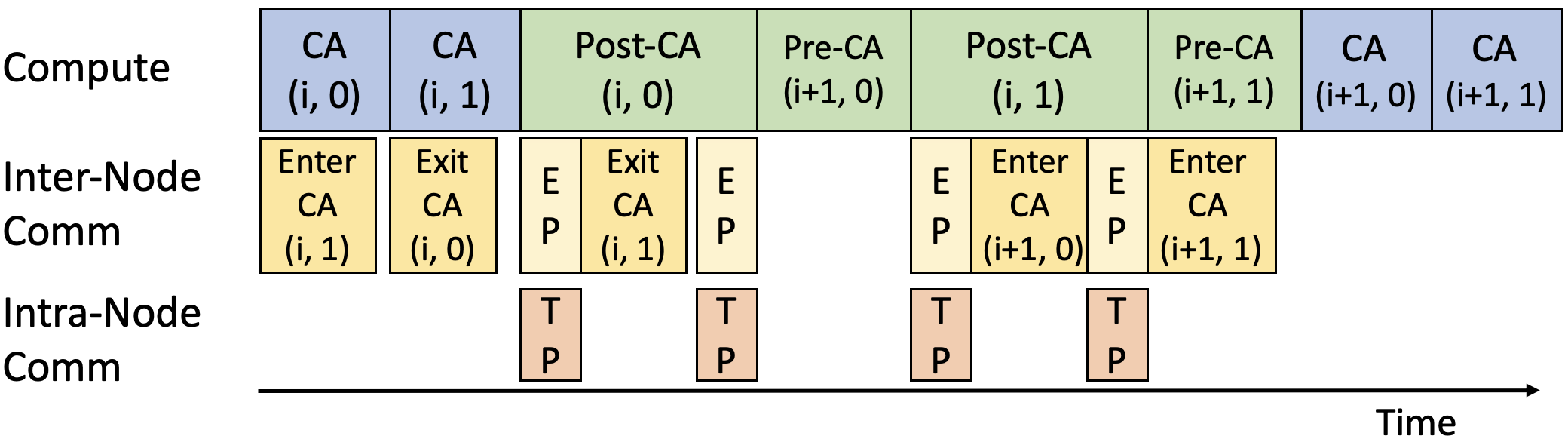}
    \caption{Ping-Pong computation and communication for inplace attention server. CA means core attention. (i, 0) means the Ping part of layer i, (i, 1) means the Pong part of layer i.}
    \label{fig:ping-pong}
\end{figure}

In CAD, core attention is computed by a pool of attention servers. 
More formally, we use \emph{core attention task} to describe attention server's workload. 
A core attention task (\texttt{CA-}task), denoted by $t$, 
is defined as the core attention computation of a query shard $q(t)$ and its context's Key-Value shard $kv(t)=\text{context}(q(t))$. 
Assume a document is split into multiple non-overlapping shards $q_1,q_2,\dots,q_n$, the whole document's core attention result is the collection of the corresponding $t_1,t_2,\dots,t_n$.

As illustrated in \autoref{fig:caaas-workflow}, 
in a group of GPUs processing multiple documents, 
each GPU can function as an independent attention server. 
For a batch of documents $B=\{d_0,d_1\dots d_n\}$, after being processed by context-independent layers, they are split into \texttt{CA-}tasks: $T=\{t^{d_0}_0,t^{d_0}_1\dots t^{d_1}_0,t^{d_1}_1\dots\}$. 
Each \texttt{CA-}task is assigned to an attention server. 
Assume a server $s$ is assigned a set of tasks, $T_s\subseteq T$. 
It first receives all inputs of the \texttt{CA-}tasks it is assigned:
$\text{Receive Tensor}=\{q(t),kv(t)|t \in T_s\}$. 
Upon receiving all input tensors, due to the divisibility discussed in \S\ref{sec:CA-disagg}, 
the server can batch all \texttt{CA-}tasks and executes within a single kernel (e.g. via a FlashAttention call). 
After that, the output of each \texttt{CA-}task is sent back to the originating GPU that handles next context-independent layers.

A central scheduler, running on the CPU, determines the sharding strategy: how to shard and generate all \texttt{CA-}tasks, then assigns each \texttt{CA-}task to an attention server, so the scheduler outputs: 
$$T,\{T_s|s\in \text{CA Servers}\} = \text{Scheduler}(B,\text{CA Servers}).$$
We detail the scheduling algorithm in \S\ref{sec:as-algo}. 
While GPUs processing the current batch, the scheduler prefetches documents for the upcoming batch. 
Using pre-computed profiling data, the scheduler estimates the computational cost for each document and potential \texttt{CA-}tasks, 
and generates a sharding and assignment plan for the batch.

Incorporating CAD into an end-to-end training system involves several key design choices, which we describe next.

\noindent \textbf{In-place attention server.} 
One straightforward design of attention server is to allocate a dedicated pool of GPUs solely for the CA computation. Although the design is feasible, we find that this design leads to significant memory underutilization for long-context training. 
As shown in \autoref{fig:mem-breakdown}, the FFN layers account for the majority of memory consumption due to their large hidden states; conversely, the core attention is stateless.
Therefore, dedicating a separate group of GPUs to core attention would leave their memory largely unused, while the GPUs processing context-independent layers would remain memory-constrained.
In-place attention server solves this problem by allowing each GPU to periodically switch its role between computing context-independent layer and acting as an attention server. By doing so, it achieves both high memory utilization and balanced compute across GPUs.

\noindent \textbf{Ping-Pong execution.} 
To hide communication overhead, we employ a ping-pong execution schedule. 
Figure~\ref{fig:ping-pong} illustrates the execution flow on a single GPU. 
We fuse the post-CA computation of the previous transformer layer with the pre-CA computation of the current layer. 
This fusion is possible because both sets of operations are context-independent. Each input microbatch is divided into two smaller nano-batches, ``Ping'' and ``Pong'', of the same number of tokens. 
The execution of these two nano-batches is interleaved, allowing the communication of one to overlap with the computation of the other. 
Furthermore, we overlap the intra-node communication for Tensor Parallel (typically over NVLink) with the inter-node communication caused by core attention disaggregation (typically over InfiniBand).

\noindent \textbf{Pipeline parallelism support.}
\begin{figure}
    \centering
    \includegraphics[width=1.0\linewidth]{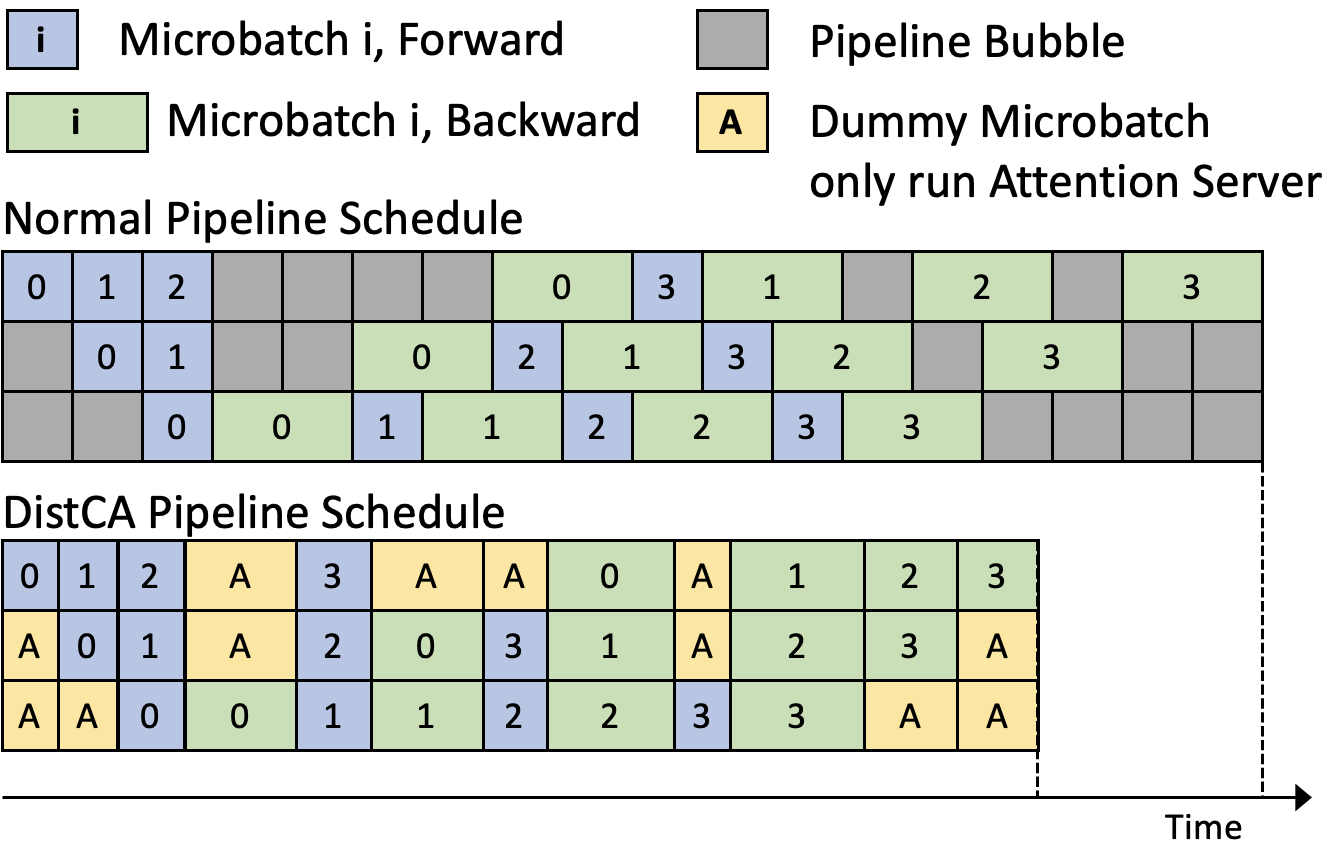}
    \vskip -0.3em
    \caption{Pipeline Parallel Schedule for normal 1F1B and disaggregated attention.}
    \label{fig:pipeline}
    \vskip -1.5em
\end{figure}
CAD naturally integrates with DP and TP, and replaces CP. We discuss how to integrate CAD with pipeline parallelism. In PP (\autoref{fig:pipeline}), a training iteration is divided into multiple logical ticks. During each tick, different pipeline stages process different microbatches concurrently. 
For CA, as it has no weights, CA tasks from different PP stages are indistinguishable from CA tasks across microbatches in DP, both of which will be scheduled and balanced across attention servers. For context-independent layers (excluding CA), since microbatches across all PP stages contain the same number of tokens, their corresponding computations will have identical -- hence balanced workload.

To prevent device from idling when switching roles between attention server and computing context-independent layers, we adjust the schedule so that all stages perform the same phase within a tick -- either all forward or all backward. We realize this by logically deferring selected backward microbatches to the pipeline bubbles at the end of the schedule, as illustrated in \autoref{fig:pipeline}, without increasing the number of ticks per iteration.
Additionally, during the pipeline warm-up and drain-down phases, some stages are inevitably idle; we repurpose these idle GPUs time for attention servers to running CA tasks. This integration is compatible with 1F1B, the interleaved 1F1B, and other widely-adopted schedules.

\subsection{Communication-Aware Greedy Scheduling} 
\label{sec:as-algo}

The scheduler must balance two competing goals: load balance and communication efficiency. Partitioning and rebatching CA tasks improves balance but might incurs transfers of $\mathbf{Q, K, V}$. We therefore solve a constrained optimization problem: (1) minimize load imbalance across attention servers (measured in FLOPs), while (2) minimizing communication volume (measured in bytes).

\noindent \textbf{Profiler.} 
To estimate the cost of a shard's \texttt{CA-}task, 
we build a profiler that benchmarks CA over a grid of query and key-value lengths $kv$. For each grid point we record ground-truth latency and throughput. 
Given a \texttt{CA-}task $t$, we predict its execution time by bilinear interpolation over the four nearest grid points. 
If $t$ lies in the saturation region  (i.e., the kernel is at peak throughput), we derive execution time from the max measured throughput instead.

\noindent \textbf{Scheduling units.}
To formally describe our scheduling algorithm, we introduce \emph{Item} as either a complete document or a shard of a document. The shard can be of any size, as long as it is a multiple of the attention kernel implementation's block size (128 in our case). Each Item already resides on the device that computes its context-independent layers. An Item's CA computation exactly maps to a \texttt{CA-}task. 

The scheduler input is a batch of Items $B$ and the number of attention servers $n$ (we assume each attention server is identical). 
The scheduler decides if an Item should be split into smaller Items, and which attention server each Item is assigned to. 
For PP, the scheduler is called for each logical tick; otherwise, the scheduler is called once per microbatch.

The scheduling algorithm unfolds in the following steps:

\textbf{1. Determine target load.} 
First, the scheduler computes the ideal per-server load ($\bar{F}$) by summing the total FLOPs of all Items divided $n$. With $\bar{F}$, attention servers are partitioned into surplus (load $> \bar{F})$ and deficit (load $ < \bar{F}$), and sorted by descending deficit.

\textbf{2. Iterating through deficit servers for migration.} 
For each deficit destination $d$ in order, we attempt to migrate from surplus sources to close $d$'s gap.  To find the most efficient Item to migrate for $d$, the scheduler evaluates each Item candidate using a \emph{cost-benefit heuristic}. 
First, for each Item, its maximum number of FLOPs to be migrated, denoted as $\Delta F_{\max}$, is determined  
by the Item's own FLOPs ($F_{\text{Item}}$), the sender's surplus ($S_{\text{source}}$), and the recipient's deficit ($D_{\text{destination}}$): $\Delta F_{\max} = \min(F_{\text{Item}}, S_{\text{source}}, D_{\text{destination}})$. To transfer each shard, there is a communication cost. We choose the shard whose FLOPs equal $\Delta F_{\max}$ but with the minimal communication. We also estimate the communication cost $V_{\text{comm}}$ associated with this migration (see \autoref{appendix:sec-comm-overhead-fn}). With $\Delta F_{\max}$ and $V_{\text{comm}}$, the scheduler calculates a \emph{priority score} for each candidate Item, defined as the communication cost per unit of computation transferred: $E = \frac{\Delta F_{\max}}{V_{\text{comm}}}$. A higher $E$ signifies a more efficient migration. 
The scheduler selects the Item with the highest score. 

Based on the calculated $\Delta F_{\max}$, the selected Item is either 
 entirely migrated if $\Delta F_{\max} = F_{\text{Item}}$, or 
spitted into two sub-Items, if $\Delta F_{\max} < F_{\text{Item}}$. 
The newly created sub-Item with $\Delta F_{\max}$ flops is dispatched to the destination attention server, while the remainder is retained by the source server.

\noindent \textbf{3. Termination Condition}
The scheduler dynamically balances the workload until when the load on each server is within $\epsilon \bar{F}$ (tolerance $epsilon$), or when remaing moves fail to improve $E$ beyond a small threshold.
In this way, the scheduler ensures the system-wise load balance while avoids unnecessary communication caused by insignificant migration.

\section{Implementations}
We implemented \sys with 2K lines of Python code. 
For efficient dispatching of attention server's input and output, we implemented an All-to-All communication kernel following the idea of~\cite{lei2025flash}, with another 1K CUDA and C++ code. The communication utilizes NVSHMEM~\cite{nvshmem}. 

Since \sys only changes the logic of attention computation, 
we integrate \sys to Megatron-LM~\cite{megatron-lm} 
to reuse its efficient implementation for token-independent layers, model architecture, 4D parallelization, and end-to-end training pipeline. 
The integration takes 1k lines of Python code for custom attention layer, ping-pong computation, and pipeline parallelism.

\section{Experiments}
\label{sec:experiments}

\subsection{Setup}

\begin{table}[t]
    \centering
    \small 
    \caption{Experiment model configurations. ``Hidden'' is the hidden dimension size, ``\#Head'' is the number of attention heads, and ``Head Size'' is the per-head dimension.}
    \vskip 0.2em
    \begin{tabular}{c|c|c|c|c|c}
        \hline
        Name & \#Layer & Hidden & \#Head & Hdim & GQA \\
        \hline
        Llama-3-8B  & 32  & 4096  & 32  & 128 & 8 \\
        Llama-34B & 48  & 8192  & 64  & 128 & 16 \\
        \hline
    \end{tabular}
    \label{tab:exp-models}
\end{table}
\noindent \textbf{Model and Hardware.} 
We test \sys on LLaMA 8B and LLaMA 34B. 
The model configurations are in \autoref{tab:exp-models}. 
All experiments are running with NVIDIA DGX H200 nodes. 
Each node has 8$\times$ 140GB H200 GPUs.

\noindent \textbf{Parallelization.}
As discussed in \S\ref{sec:background-llm-arch-and-parallel}, TP offers load balance among ranks, 
and has negligible communication overhead when the communication is within a device. 
So we fix TP=8 in our experiments. 
For PP, we grid search best configurations that avoid out-of-memory(OOM).

After determining TP and PP degrees, we grid search DP and CP degree for the baseline methods. 
For \sys, documents are placed sequentially: each device computes a fixed number of tokens for context-independent layers. 
If a device reaches its token threshold before a document is fully placed, 
the remaining portion of that document is put to the next device.

\noindent \textbf{Input data.} 
For each experiment, given a number of tokens per batch, 
we sample 30 batches from an input distribution 
and report the average throughput. 
Our experiments use two (synthetic) distributions: 
(1) Pretrain with upsampled long context documents (``Pretrain''). 
Following the common practice~\cite{fu2024data}, 
we upsample long documents in a pretrain data distribution by randomly filtering out documents shorter than a threshold. 
(2) ProLong~\cite{gao2024train}: a public dataset specific for long context training. 
This work shows that a mixture of long and short documents brings the best performance. 
Compared to ``Pretrain'', ``ProLong'' has a higher percentage of long documents.

We vary the maximum number of tokens (``MaxDocLen'') to evaluate the system's performance with different context window size. 
The total number of tokens is determined by the memory capacity. 
In all test cases, the baseline goes out of memory before \sys, 
and the total number of tokens for all systems are set to that value.

\noindent \textbf{Baseline.} We compare against WLB-LLM~\cite{wang2025wlb} (``WLB-ideal'') as our primary baseline. 
Since no official implementation of the full WLB-LLM system is available, we reimplement its method with our system. 
To reproduce the adaptive CP sharding policy in WLB-LLM, 
we sweep the DP-CP degree and report the throughput of the best-performing configuration. 
For the variable-length document packing, 
we do not implement the deferred execution mechanism (Algorithm 1 in~\cite{wang2025wlb}), 
as it alters the training dynamics; we leave exploring its integration to future work.

\subsection{End-to-End Experiment}

\begin{table}[t]
\centering
\footnotesize
\caption{3D Training Configurations.}
\vspace{3pt}
\setlength{\tabcolsep}{4pt}
\renewcommand{\arraystretch}{1.0}
\begin{tabular}{lcll}
\toprule
\textbf{Model} & \textbf{MaxDocLen} & \textbf{Batch Size} & \textbf{\#GPU} \\
\midrule
Llama-8B  & 128K & 8, 16, 32  & 64, 128, 256 \\
Llama-8B  & 256K & 4, 8, 16   & 64, 128, 256 \\
Llama-8B  & 512K & 2, 4, 8    & 64, 128, 256 \\
\hline
\addlinespace[2pt]
Llama-34B & 128K & 4, 8, 16   & 64, 128, 256 \\
Llama-34B & 256K & 2, 4, 8    & 64, 128, 256 \\
Llama-34B & 512K & 2, 4, 8    & 64, 128, 256 \\
\bottomrule
\end{tabular}
\label{tab:3d-config}
\end{table}

\begin{figure}
    \centering
    \includegraphics[width=0.95\linewidth]{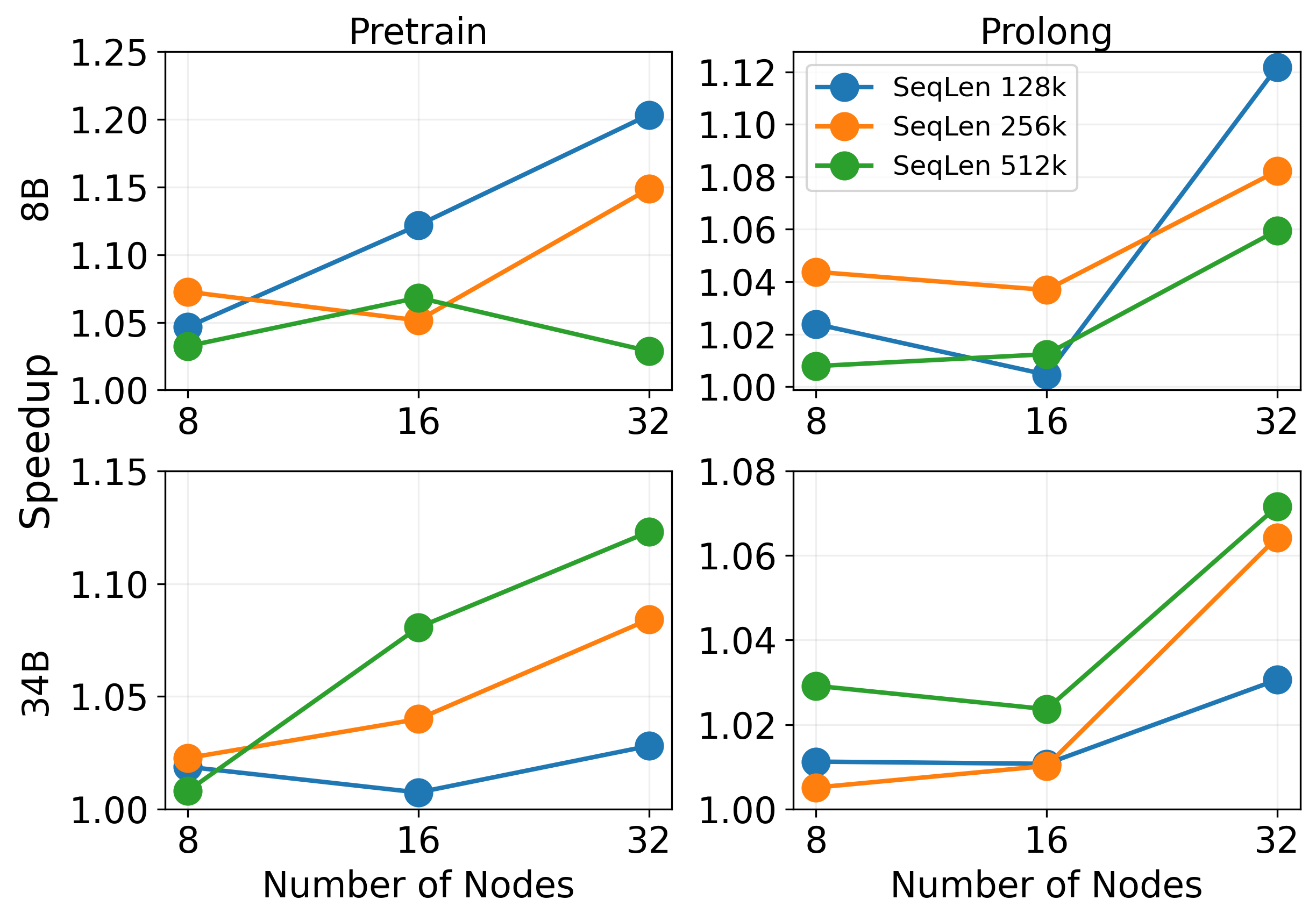}
    \caption{3D Parallel (no PP) experiment. Speedup is defined as the average duration of the runs of WLB-LLM over \sys.}
    \label{fig:dpcp-2x2-comparison}
\end{figure}

\noindent \textbf{3D parallelism (w/o PP).} We first show the performance without PP. Table~\ref{tab:3d-config} shows the ``MaxDocLen", the batch size, and number of GPUs for each experiment.

The result is shown in Figure~\ref{fig:dpcp-2x2-comparison}. \sys consistently outperforms ``WLB-ideal'', 
showing 1.07-1.20x speedup in Pretrain, and 1.05-1.12x speedup in Prolong dataset, and has a better scaling behavior. 
The primary reason is that, 
``WLB-ideal'' suffers from either a higher context parallelism overhead (higher CP) or a more challenging scheduling demands (higher DP), as discussed in \S\ref{sec:imbalance-motivating-example}. 
In contrast, \sys only dispatches a subset of tokens, and hides the communication overhead by ping-pong execution. 
The flexible token-level scheduling space also enables a better load balance. 
Moreover, as discussed in \S\ref{sec:imbalance-motivating-example}, two conflicting factors constrain ``WLB-ideal'' on the memory side.
One one hand, variable-size data chunks across DP ranks exacerbate memory imbalance. 
On the other hand, CP introduces substantial all-gather communication.

We also observe that \sys has a higher speedup on ``Pretrain'' than ``ProLong''. 
This is mainly because ``Pretrain'' contains a higher proportion of short documents, 
and is more challenging for ``WLB-ideal’’ to balance the workload.

With 34B model, \sys achieves greater speedups at higher ``MaxDocLen''. This is because a larger ``MaxDocLen'' leads to a more diverse document length distribution, making it increasingly difficult for ``WLB-ideal'' to balance the workload effectively.

Interestingly, with 8B model, \sys achieves greater speedups at lower ``MaxDocLen''. 
This is because we use the same number of nodes to train the 8B model as the 34B model. 
As a result, the total number of tokens per batch is larger, increasing the likelihood of having multiple documents with similar lengths. 
In this setting, the all-gather overhead becomes more significant: a smaller ``MaxDocLen'' leads to lower FLOPs per token in the core attention computation, while the all-gather cost remains the same because the total number of tokens is the same.

\begin{table}[t]
\centering
\footnotesize
\caption{4D Parallel Training Configurations.}
\vspace{3pt}
\setlength{\tabcolsep}{4pt}
\renewcommand{\arraystretch}{1.0}
\begin{tabular}{lcll}
\toprule
\textbf{Model} & \textbf{MaxDocLen} & \textbf{Batch Size} & \textbf{\#GPU} \\
\midrule
Llama-8B  & 128K  & 32, 64, 128 & 64, 128, 256 \\
Llama-8B  & 256K  & 16, 32, 32  & 64, 128, 256 \\
Llama-8B  & 512K  & 8, 8, 16    & 64, 128, 256 \\
\hline
\addlinespace[2pt]
Llama-34B & 128K  & 32, 64, 128 & 128, 256, 512 \\
Llama-34B & 256K  & 16, 32, 32  & 128, 256, 512 \\
Llama-34B & 384K  & 8, 8, 16    & 128, 256, 512 \\
\bottomrule
\end{tabular}
\label{tab:4d-config}
\end{table}

\begin{figure}
    \centering
    \includegraphics[width=0.95\linewidth]{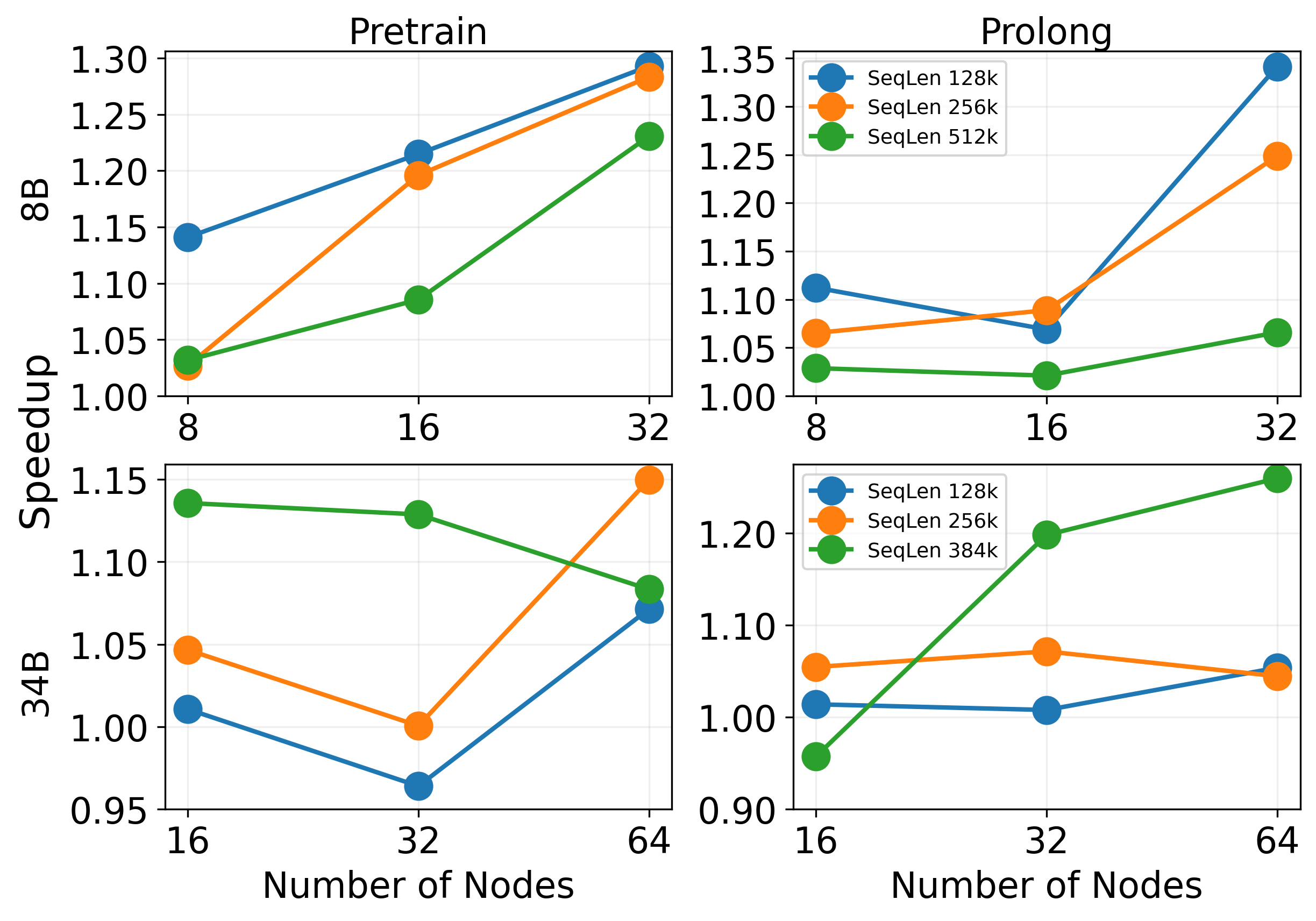}
    \caption{4D Parallel (with PP) experiment. Speedup is defined as the average duration of the runs of WLB-LLM over \sys.
    }
    \label{fig:pp-2x2-comparison}
\end{figure}

\noindent \textbf{4D parallelism.} We now present the performance results under full 4D parallelism. Table~\ref{tab:4d-config} shows the ``MaxDocLen", the batch size and number of GPUs for each experiment. 
In each setting, we sweep both \sys and WLB across all possible configurations, using a sufficiently large batch size for the given model and workload. 
To ensure a fair comparison for WLB in 34B model, we fix the number of GPUs only for the 34B model to 16, 32, and 64, thereby enlarging the pipeline parallelism search space.

The results are shown in Figure~\ref{fig:pp-2x2-comparison}. \sys generally outperforms ``WLB-ideal’’ and demonstrates more favorable scaling characteristics over the set of different configurations. 
For the 8B model (top 2 figures in Figure~\ref{fig:pp-2x2-comparison}), \sys achieves 1.15x-1.30x speedup on the Pretrain dataset and 1.10x-1.35x speedup on the ProLong dataset across different configurations. 
In addition to the advantages described with the 3D parallel (non PP) experiments, 
\sys further balances computation across pipeline stages, and repurpose idle stages during warmup and drain-out phases as core attention servers. 
In contrast, WLB-LLM struggles to find effective configurations, often running out of memory at high CP or DP degrees, and experiencing amplified load imbalance caused by pipeline parallelism, which further skews attention computation across stages.

For the 34B model, \sys also shows positive speedup across most configurations (16, 32, and 64 GPUs), achieving up to 1.15x speedup on Pretrain and 1.25x speedup on ProLong. The performance gap generally widens as the maximum document length increases, since larger input diversity exacerbates WLB’s load imbalance. 

Despite these gains, we observe that memory fragmentation creates some runtime overhead that limits \sys’s performance in the 34B 4D-parallel experiments. 
Because Core Attention handles requests with varying tensor shapes at each micro-batch, the allocator repeatedly creates and releases differently sized memory blocks, causing fragmentation and frequent PyTorch garbage collection. 
The resulting CPU overhead delays GPU kernel launches and degrades overall performance. 
We plan to address this issue in future work with static memory allocation and CUDA Graphs.

\subsection{Ablation study}
To study the effects of different components in \sys, we designed several ablation studies:

\begin{figure}
    \centering
    \includegraphics[width=\linewidth]{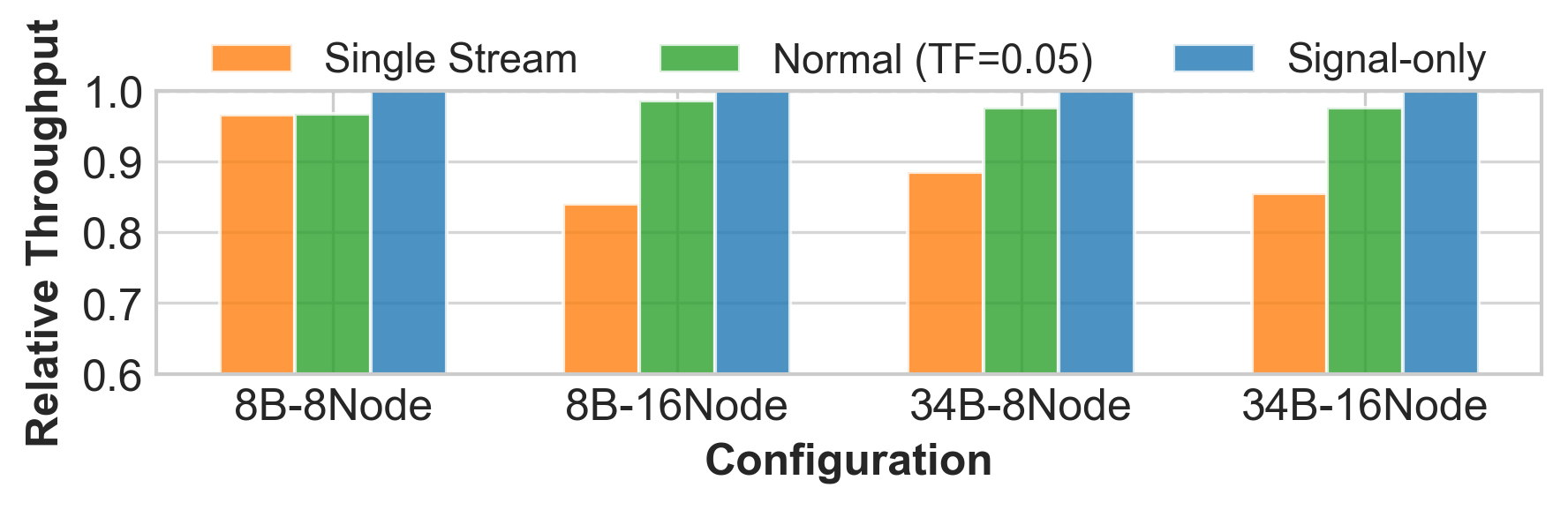}
    \caption{Throughput for different communication patterns.}
    \label{fig:system-overhead}
\end{figure}

\noindent \textbf{System overhead.} The primary system overhead in \sys lies in the communication volume 
between a token's attention server and the device for its context-independent layers. 
We designed two baselines to illustrate the overhead:

1) Signal communication (``Signal''). This reduces each communication volume to 1 byte, 
meaning the synchronization overhead now only reflects computation imbalance.

2) Remove Ping-Pong execution (``Single Stream''). This makes the communication on the same GPU stream as computation, removing the optimization to overlap the two parts.

% We tested on both 8B and 34B models on 8 and 16 nodes using the Pretrain distribution, fixing the total number of tokens to saturate the compute. 
% Figure~\ref{fig:system-overhead} shows our experiment on both 8B and 34B models on 8 and 16 nodes using the Pretrain distribution, fixing the total number of tokens to saturate the compute.
Figure~\ref{fig:system-overhead} reports results on 8B and 34B models across 8 and 16 nodes using the Pretrain distribution, with the total number of tokens fixed to saturate compute.
\sys achieves nearly the same latency as ``Signal'', indicating that communication is almost fully overlapped with computation, and therefore eliminating the 10-17\% higher latency that would otherwise incur by ``Single Stream’’ incurs .The only exception is the 8B model on 8 nodes, where the compute workload is too small to fully hide communication.
% Figure~\ref{fig:system-overhead} shows that
% \sys has almost the same latency as ``Signal'', while ``Single Stream'' has a 10-17\% higher latency. 
% This means that in \sys, the communication almost perfectly overlaps with the computation, 
% reducing what would otherwise be a 10-17\% overhead. 
% The only exception is with the 8B model on 8 nodes, where the computation is relatively small, making it harder to fully hide communication.

\begin{figure}
    \centering
    \includegraphics[width=\linewidth]{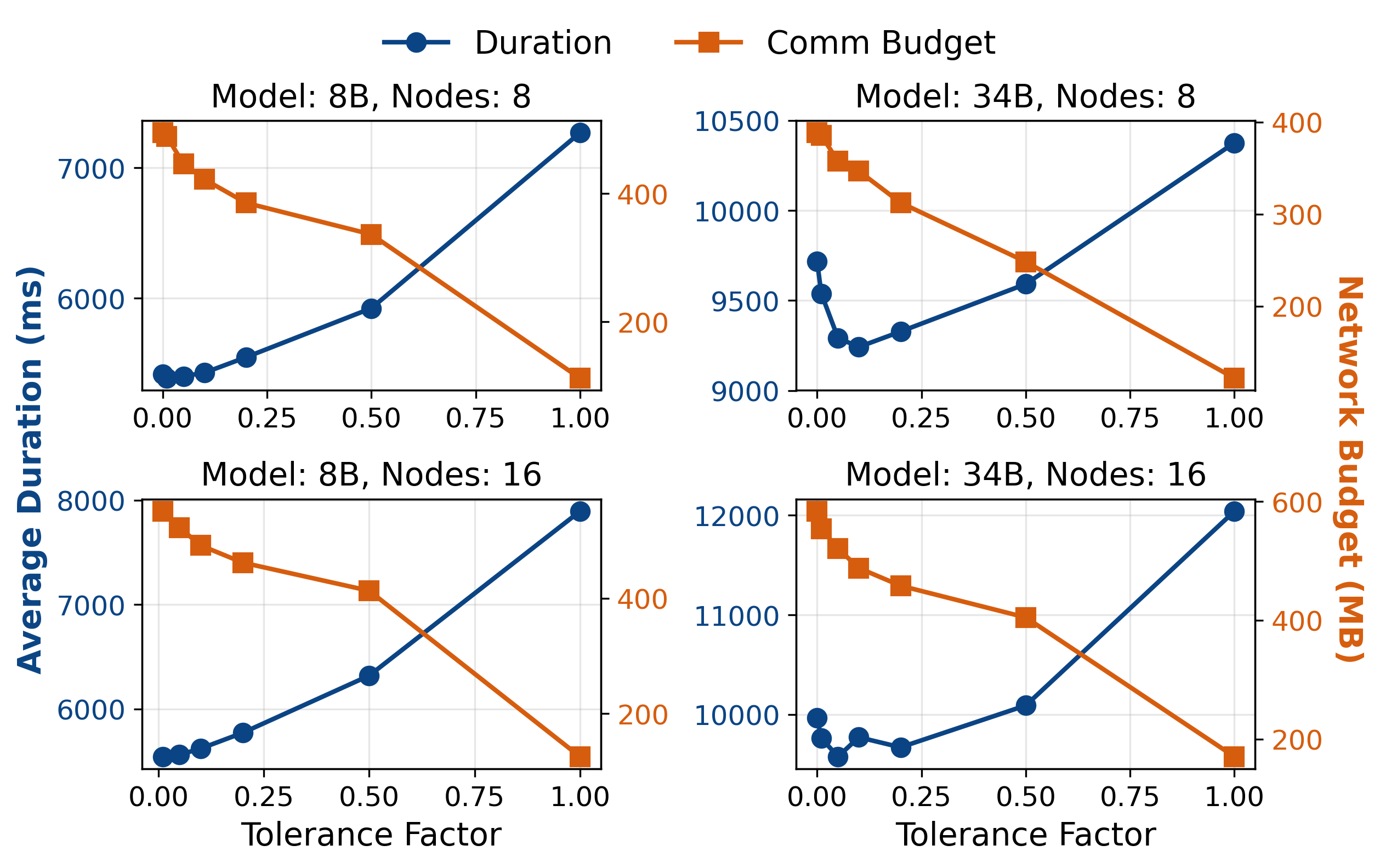}
    \caption{Impact of the compute imbalance tolerance factor.}
    \label{fig:ablation.tolerance_factor}
\end{figure}

\noindent \textbf{Hyper-parameter in scheduler.} 
% Our scheduler introduces a tolerance factor, which shows a trade-off between 
% a more balanced CA computation 
% and a lower communication volume. 
The scheduler introduces a tolerance factor that trades off CA load balance against communication volume to minimize latency. 
% We evaluate this trade-off by measuring latency and communication volume across different tolerance factors, using 8B and 34B models on 8 and 16 nodes, with 1M and 512K total tokens respectively, and a 128K maximum document length.
% In this experiment, we quantitatively analyze such a tradeoff by showing the latency and communication volume with different tolerance factor. 
% We use both 8B and 34B models, on 8 and 16 nodes, with 1M and 512K total tokens respectively, and 128K maximum document length.
% Figure~\ref{fig:ablation.tolerance_factor} shows the result.
Figure~\ref{fig:ablation.tolerance_factor} shows this trade-off by evaluating 8B and 34B models on 8 and 16 nodes under the Pretrain distribution, using 1M and 512K total tokens respectively with a maximum document length of 128K.
For the 8B model, latency remains largely unchanged when the tolerance factor is between 0 and 0.20 as the communication can still be fully overlapped with computation. For the 34B model, however, setting the tolerance factor below 0.10 is too restrictive -- communication volume increases and can no longer be hidden, causing higher latency. Conversely, when the tolerance factor is too large, latency roughly linearly rises again due to load imbalance.
% At the other extreme, increasing the tolerance factor beyond this point causes latency to rise again approximately linearly, as the resulting load imbalance becomes increasingly significant. 

Figure~\ref{fig:ablation.tolerance_factor} also plots the relation between communication size and imbalance tolerance factor. 
In most cases, tuning tolerance factor from 0 to 0.15 decreases the memory requirement by 20-25\% while 
leaving average duration staying almost the same or -- in some cases as shown in 34B, 8 nodes -- better. 
A tolerance factor beyond this point will significantly increase the iteration latency, while leaving the communication size relatively stable.

\section{Related Works}

\noindent \textbf{Load imbalance in long-context training.}
\citet{lin2025understanding} identifies the document length skew across microbatches as a primary cause of slowdowns in today's LLM training systems.
Several strategies have been proposed to mitigate this issue.
FlexSP~\cite{wang2025flexsp} introduces dynamic context parallelism. It uses an integer linear programming (ILP) solver to dynamically choose CP degree for device subgroups and to assign documents to these subgroups. However, the ILP is NP-hard, which limits scalability; besides, it only targets a narrower setting (FSDP). 
Several other approaches also attempt to mitigate the CP overhead either by dynamic CP degree for each document~\cite{byteps} or by overlapping communication with attention computation~\citep{liu2023ring,zeppelin}. Compared to these variants of CP, CAD achieves a more precise load balance by a token granularity variable length sharding. CAD also reduces the communication at scale by not only preventing sharding short document, but sharding long documents aware of communication efficiency.

% WLB-LLM
In addition to per-document CP, WLB-LLM~\cite{wang2025wlb} introduces variable-size data chunk, which uses an imbalanced MLP to compensate the workload imbalance from attention. Because activation memory is tied to MLP computation, equalizing total FLOPs with this method leads to memory imbalance. 
As the context length increases, the fraction of MLP computation shrinks, making it harder to offset attention imbalance.

\noindent \textbf{Model disaggregation.} 
Disaggregation, most commonly prefill-decode disaggregation, has been widely adopted in LLM inference~\cite{distserve,splitwise,tetriinfer,qin2025mooncake}. 
The most closely related efforts MegaScale-Infer~\cite{zhu2025megascale} and concurrent work~\cite{wang2025step} further disaggregate attention and FFN onto separate physical devices in inference. These work primarily targets mixture-of-expert (MoE) models as the per-layer transfer caused by disaggregation can be merged with the token routing communication inherent to MoE, thereby avoiding additional latency; for dense models, they would incurs extra overhead. In contrast, our work focuses on \emph{language model training}, which is throughput-oriented, and demonstrates benefits independent of model architectures.

\section{Limitations}

\sys implements in-place attention servers to maintain memory utilization (\S\ref{sec:system-features}). If memory demand is satisfied, dedicating more GPUs to attention (without scaling those for others) could further reduce compute time while preserving load balance and low communication overhead. We also believe that \sys could enable better fault tolerance and performance isolation with a dedicated pool, which we leave to future work.

Our scheduler restricts each \texttt{CA-}task to a $\mathbf{Q}$ shard with the full $\mathbf{K,V}$ context. Allowing a \texttt{CA-}task to use a $\mathbf{Q}$ shard with only a sub‑range of its  $\mathbf{K,V}$  context would add flexibility. In addition, when estimating communication, the current model pessimistically assumes all tokens are transferred and ignores $\mathbf{K,V}$ already resident on the destination; this can overestimate bytes and yield non‑minimal transfers.

\section{Conclusion}
This paper presents core attention disaggregation, a new architecture for large language model training that separates the core attention module from the rest of the model to enable independent scaling and scheduling.
We observe that core attention is stateless and composable at token granularity, enabling efficient disaggregation with minimal communication overhead. 
We implement core attention disaggregation in \sys. \sys features a workload-aware scheduler to balance computation while minimizing communication, and a ping-pong execution scheme to hide dispatch latency. End-to-end evaluation shows that \sys improves throughput by up to 1.35x speedup over state-of-the-art training systems, with increasing advantages at larger scales.


\bibliographystyle{mlsys2025}
\bibliography{reference}

%%%%%%%%%%%%%%%%%%%%%%%%%%%%%%%%%%%%%%%%%%%%%%%%%%%%%%%%%%%%%%%%%%%%%%%%%%%%%%%
%%%%%%%%%%%%%%%%%%%%%%%%%%%%%%%%%%%%%%%%%%%%%%%%%%%%%%%%%%%%%%%%%%%%%%%%%%%%%%%
% SUPPLEMENTAL CONTENT AS APPENDIX AFTER REFERENCES
%%%%%%%%%%%%%%%%%%%%%%%%%%%%%%%%%%%%%%%%%%%%%%%%%%%%%%%%%%%%%%%%%%%%%%%%%%%%%%%
%%%%%%%%%%%%%%%%%%%%%%%%%%%%%%%%%%%%%%%%%%%%%%%%%%%%%%%%%%%%%%%%%%%%%%%%%%%%%%%
\appendix

\section{Upper Bound for Core Attention Server max partition size}
\label{appendix:upper-bound-ca-cp-degree}
\begin{table}[ht]
    \centering
    \caption{Llama-34B configuration}
    \begin{tabular}{c|c|c}
         hidden ($h$) & key-value hidden ($h_{kv}$) & intermeidate ($i$) \\
         8192 & 2048 & 22016
    \end{tabular}
    \label{tab:llama-config}
\end{table}

Let a document of length $l$ be divided into $s$ shards. 
We denote the hidden size of query and key-value tensor as $h_q$ and $h_{kv}$. 
The query states for the entire document must be distributed, 
resulting in a total communication volume of $l\cdot h_q$. 
Besides, the first key-value shard serves as context of all query shards (0 to $s-1$), 
the second serves the context of query shards 1 to $s-1$...
Assuming the document is evenly sharded, the total communication volume for key-value states is:
$h_{kv}\cdot (l/s \cdot s + l/s\cdot (s-1)+\dots) = (s+1)lh_{kv}/2$. 

Let the network bandwidth be $B$, and the time to compute a token's Context-Independent layer be $t$, 
When communication is fully overlapped with computation, there should be $t\cdot l\ge l\cdot (h_q + h_{kv}(s+1)/2)/B$. 
Rearranging the terms gives an upper bound on the number of shards: $s \le 2(tB - h_q)/h_{kv} - 1$.

We use Llama-34B as the example to analyze the theoretical max number of partition. 
Its configuration is listed in ~\autoref{tab:llama-config}. 
We assume the InfiniBand bandwidth is 50GB/s, 
and the MFU for Context Independent layer is 50\% of the H200 node (990TFLOPs for FP16). 

The total flops to compute a token's Context Independent layer is:
\begin{align*}
    2 \cdot (h \cdot h \cdot 2 + h\cdot h_{kv} + h\cdot i\cdot 3) &= 2h(2h + h_{kv}+3i) \\
    & = 1320\cdot 2^{20}\text{ flops}
\end{align*}

Here the first factor 2 is because 
each pair of elements in matrix multiplication has an add operation and a multiplication operation. 
The Query and Output tensors each has a mapping from a vector of hidden size to another also of hidden size; 
the Key and Value tensors each has a mapping from hidden size to key-value hidden size; 
the Feed-Forward Layer applies a Gated 2-layer MLP, so it has 3 mapping from hidden size to FFN intermediate size, or reversely from FFN intermediate size to hidden size. 

As a result, the time to compute these layers are:
\begin{equation}
    t = 1320\cdot 2^{20}\text{ flops} / (50\% \cdot 990\text{ TFLOPs}) \approx 2.796\cdot 10^{-6}\text{s}
\end{equation}

Taking this into the formulation $s\le 2(tB - h_q) / h_{kv} - 1$, we have:
\begin{align*}
    s & \le 2(tB - h_q) / h_{kv} - 1 \\
    & = 2(2.796\cdot 10^{-6}\text{s} \cdot 50\text{GB/s} - 16\text{KB}) / 4\text{KB} - 1 \\
    & \approx 31
\end{align*}

The computation time for context-independent layers, $t$, scales quadratically with the hidden size $h_q$. 
Therefore, for larger models, this upper bound on $s$ even increases.

\section{Communication overhead function}
\label{appendix:sec-comm-overhead-fn}

This section discusses the formulation of communication cost function $v(\cdot)$. 

For a shard of $n_q$ query tokens and $n_{kv}$ key-value tokens, there is:
\begin{align*}
    v(\Delta F_{\max}, L_q, L_{kv}, \text{size}_q, \text{size}_{kv}) & = \min_{n_q,n_{kv}} \text{Comm}(n_q,n_{kv}) \\
    0 < n_q & \le L_q \\
    n_q + L_{kv} - L_q \le n_{kv} & \le L_{kv}   \\
    \frac{n_q (2n_{kv} - n_q)}{L_q (2L_{kv} - L_q)} & = \frac{\Delta F_{\max}}{F_{\text{Item}}} \\
    \text{Comm}(n_q,n_{kv}) & = n_q\cdot \text{size}_q + n_{kv}\cdot \text{size}_{kv}
\end{align*}
Let $\alpha=\frac{\Delta F_{\max}}{F_{\text{Item}}}, \beta=\frac{\text{size}_{kv}}{\text{size}_q}$, there is:
$$\text{Comm}(n_q,n_{kv}) = \text{size}_{q} (n_q (1+\frac{1}{2}\beta) + \frac{\alpha\beta L_q(2L_{kv} - L_q)}{2n_q})$$
As a result, selecting a shard of $n_q^{\text{opt}}=\sqrt{\alpha\beta L_q(2L_{kv} - L_q) / (\beta + 2)}$ results in the minimal communication: $v(\cdot) = \text{size}_{q} \cdot \sqrt{\alpha\beta(\beta + 2) L_q(2L_{kv} - L_q)}$. 
However, if there is $n_q^{\text{opt}}> L_q$ or the corresponding $n_{kv}^{\text{opt}} < n_q + L_{kv} - L_q$, there is the optimal configuration corresponds to the upper bound of $n_q^{\max} = \sqrt{(L_q - L_{kv})^2 + \alpha L_q(2L_{kv} - L_q)}$, and the minimal communication is the corresponding value.

In practice, we notice that different shards have a diverged MFU. As a result, estimating the Item (or the sub-Item)'s time directly by flops is not precise. 
However, our profiling result also shows that, if we still follow the practice of the head-tail context parallel, i.e. an Item has both the first i to j tokens and the last i to j tokens, the estimation by flops is still accurate. 
As a result, we keep using the head-tail context parallel and leave a more precise modeling as a future work. 
In this way, we redefine $n_{kv}$ as the key-value tokens for the ``head'' half shard, and the communication cost is modified as: 
$$\text{Comm}(n_q,n_{kv}) = n_q\cdot \text{size}_q + (L_{\text{doc}} - (n_{kv} - n_{q}))\cdot \text{size}_{kv}$$
This is because that the first i to j tokens need the first j token's KV states, while the last i to j tokens need the $L_{\text{doc}} - i$ token's KV states. By definition, there is $i + n_q = j = n_{kv}$. 
Hence, the communication equals:
$$L_{\text{doc}}\text{size}_{kv} + \frac{1}{2}\text{size}_q \left(n_q(2+\beta) - \frac{\alpha\beta L_q(2L_{kv} - L_q)}{n_q}\right)$$
As a result, the optimal $n_q$ corresponds to the minimal possible value. 
Based on other constraints, there is
\begin{align*}
    n_q^{\min} = &L_{kv} - \sqrt{L_{kv}^2 - \alpha(2L_{kv} - L_q)L_q} \\
    v(\cdot) = & L_{doc}\text{size}_{kv} +\\& \frac{1}{2}\text{size}_q\left(n_q^{\min}(2+\beta) - \frac{\alpha\beta L_q(2L_{kv} - L_q)}{n_q^{\min}}\right)
\end{align*}

%%%%%%%%%%%%%%%%%%%%%%%%%%%%%%%%%%%%%%%%%%%%%%%%%%%%%%%%%%%%%%%%%%%%%%%%%%%%%%%
%%%%%%%%%%%%%%%%%%%%%%%%%%%%%%%%%%%%%%%%%%%%%%%%%%%%%%%%%%%%%%%%%%%%%%%%%%%%%%%

\end{document}